\documentclass[letterpaper,journal]{IEEEtran}

\usepackage{amsmath,amsfonts}
\usepackage{algorithmic}
\usepackage{algorithm}
\usepackage{array}
\usepackage[caption=false,font=normalsize,labelfont=sf,textfont=sf]{subfig}
\usepackage{textcomp}
\usepackage{stfloats}
\usepackage{url}
\usepackage{verbatim}
\usepackage{graphicx}
\usepackage[hidelinks]{hyperref}
\usepackage{cite}
\usepackage{comment}
\newcommand{\orcid}[1]{\href{https://orcid.org/#1}}
\usepackage{amssymb}

\newcommand{\vct}[1]{\mathbf{#1}}         
\newcommand{\mtr}[1]{\mathbf{#1}}         
\newcommand{\vp}{\vct{p}}
\newcommand{\vzero}{\mathbf{0}}           
\newcommand{\bone}{\mathbf{1}}
\newcommand{\transpose}{\intercal}

\newcommand{\abs}[1]{\left\lvert #1 \right\rvert}
\newcommand{\norm}[1]{\left\lVert #1 \right\rVert}

\DeclareMathOperator{\diag}{diag}

\newcommand{\calP}{\mathcal{P}}

\newcommand{\R}{\mathbb{R}}

\newcommand{\vecC}{\vct{c}}
\newcommand{\matA}{\mtr{A}}
\newcommand{\vecB}{\vct{b}}
\newcommand{\matG}{\mtr{G}}
\newcommand{\vecH}{\vct{h}}

\newcommand{\vx}{\vct{x}}
\newcommand{\vs}{\vct{s}}
\newcommand{\vxi}{\boldsymbol{\xi}}

\newcommand{\vy}{\vct{y}}
\newcommand{\vz}{\vct{z}}
\newcommand{\vw}{\vct{w}}

\newcommand{\vq}{\vct{q}}

\newcommand{\vlam}{\boldsymbol{\lambda}}

\newcommand{\vecZeta}{\boldsymbol{\zeta}}

\newcommand{\vr}{\vct{r}}

\newcommand{\vv}{\vct{v}}

\newcommand{\tres}[1]{\tilde{r}_{#1}}

\newcommand{\NatMerit}{\mathcal{M}}

\newcommand{\vncpFun}[1]{\boldsymbol{\phi}^{#1}}

\newcommand{\ncpJac}[1]{\boldsymbol{\Phi}_{#1}}
\newcommand{\pertncpJac}[1]{\tilde{\boldsymbol{\Phi}}_{#1}}

\newcommand{\rhod}{\rho_{d}}      
\newcommand{\rhoe}{\rho_{e}}      
\newcommand{\rhoi}{\rho_{i}}      
\newcommand{\rhon}{\rho_{n}}      

\newcommand{\xE}{\vx_{E}}
\newcommand{\yE}{\vy_{E}}
\newcommand{\zE}{\vz_{E}}
\newcommand{\sE}{\vs_{E}}

\newcommand{\matP}{\mtr{P}}
\newcommand{\matD}{\mtr{D}}
\newcommand{\matH}{\mtr{H}}       

\newcommand{\Id}{\mtr{I}}

\newcommand{\dx}{\Delta\vx}
\newcommand{\dy}{\Delta\vy}
\newcommand{\dz}{\Delta\vz}
\newcommand{\ds}{\Delta\vs}

\newcommand{\epsa}{\epsilon_{a}}      
\newcommand{\epsr}{\epsilon_{r}}      
\newcommand{\epsp}{\epsilon_{p}}      
\newcommand{\epsd}{\epsilon_{d}}      

\newcommand{\control}{\vct{u}}

\newcommand{\matL}{\mathbf{L}}

\newcommand{\calF}{\mathcal{F}}

\newcommand{\vecZetaE}{\vecZeta_{E}}
\newcommand{\vlamE}{\vlam_{E}}
\newcommand{\tauE}{\tau_{E}}
\newcommand{\kappaE}{\kappa_{E}}

\newif\ifpublished
\newcommand{\published}[1]{%
  \ifnum\pdfstrcmp{#1}{True}=0\relax
    \publishedtrue
  \else
    \publishedfalse
  \fi
}
\published{True} 

\begin{document}

\title{HALO-LP: A Homogeneous All-Shifted\\
Non-Interior-Point Method for Linear Programming}

\ifpublished
\author{
    Jose Rojas$^{1}$\quad
    Ana Aguilar$^{2}$\quad
    Eusebio Bugarin$^{2}$
\thanks{$^{1}$Robot Motor Intelligence (RoMI) Lab -- Heriot-Watt University, U.K.}
\thanks{$^{2}$Department of Electrical and Electronic Engineering,
Tecnol\'ogico Nacional de M\'exico/I.T. Ensenada,
Ensenada, B.C., Mexico.}
}
\else
\author{Author Names Omitted for Anonymous Review.}
\fi

\ifpublished
\markboth
{Rojas \MakeLowercase{\textit{et al.}}:
HALO-LP: A Homogeneous All-Shifted Non-Interior-Point Method for Linear Programming}
{}
\else
\markboth
{Anonymous \MakeLowercase{\textit{et al.}}:
HALO-LP: A Homogeneous All-Shifted Non-Interior-Point Method for Linear Programming}
{}
\fi



\maketitle

\begin{abstract}
This letter introduces HALO-LP, a novel homogeneous all-shifted
non-interior-point method for linear programming (LP).
The proposed method combines a homogeneous self-dual
(HSD) embedding with all-shifted nonlinear complementarity
(NCP) functions and proximal primal--dual regularization,
enabling arbitrary primal--dual warm starts without
requiring strictly interior iterates. The resulting Newton equations admit a condensation based on
a positive-definite matrix that can be efficiently factorized using Cholesky decomposition, while the HSD embedding enables primal and dual
infeasibility detection through Farkas certificates.
Numerical evaluations on perturbed LP instances derived
from the Netlib test set demonstrate
state-of-the-art warm-starting performance, outperforming
the competing solvers in the evaluated setting.
We further validate optimality and infeasibility detection
on analytically characterized LP problems.
Finally, we demonstrate HALO-LP's applicability to
robotic optimal control through sequential linear
programming (SLP) for centroidal walking and trotting
of quadrupedal robots.
\end{abstract}


\section{Introduction}
Optimization-based planning and control methods are widely
used in robotics to generate motions while enforcing dynamics,
contact, actuator, and environment constraints~\cite{mastalli2023inverse}. In particular,
model predictive control (MPC) repeatedly solves closely
related optimization problems as new state estimates and
environment information become available. Sequential linear
programming (SLP) provides an attractive approach to
nonlinear trajectory optimization by solving successive
linear approximations of the original problem~\cite{byrd2005convergence}.

These applications require efficient LP solvers capable of
exploiting previous primal--dual solutions while reliably
detecting infeasible or unbounded problems. Such capabilities
are also important in mixed-integer and generalized
disjunctive programming (GDP), where sequences of related
LP relaxations must be solved and infeasible subproblems
identified. In contact-rich robotics, these challenges
are further amplified by changing active constraints and
contact configurations.

Primal--dual interior-point methods (IPMs) are among the
most successful algorithms for large-scale linear
optimization. However, their strict interiority requirements
complicate warm-starting, as previously computed solutions
typically lie near the boundary of the feasible region~\cite{benson2007,engau2009}.
Homogeneous self-dual (HSD) embeddings provide a unified
framework for recovering optimal solutions and detecting
infeasibility~\cite{ye1994}, but conventional HSD interior-point methods
retain these interiority restrictions.

The central challenge is to preserve the optimality and
infeasibility certification capabilities of HSD methods
while removing the interiority restrictions that hinder
warm-starting. To address this challenge, we introduce
HALO-LP, which regularizes the HSD equations rather than
the original LP problem. The proposed all-shifted
formulation allows iterates to leave the positive
orthant while maintaining a well-conditioned Newton
system. At a proximal fixed point, the shifted perturbations vanish; as the smoothing parameter approaches zero, the original HSD optimality and infeasibility conditions are recovered. This provides a unified framework for warm-started linear optimization and infeasibility detection.

\section{Related Work}

\subsection{Warm-starting interior-point methods}

Warm-starting primal--dual IPMs remains challenging because previously computed solutions typically lie near the boundary of the feasible region.
Benson and Shanno~\cite{benson2007} proposed an exact
primal--dual penalty reformulation that relaxes
nonnegativity constraints, while Engau
et al.~\cite{engau2009} developed a primal--dual slack
approach for reoptimization. Gondzio and
Grothey~\cite{gondzioGrothey2008} introduced an unblocking
technique based on sensitivity analysis of Newton
directions. More recently, Gill and
Zhang~\cite{gill2024} proposed an all-shifted
primal--dual penalty--barrier method with a projected line
search to enable warm-starting in IPMs. However, these
approaches retain a barrier-based treatment of complementarity,
whereas HALO-LP permits non-interior primal--dual iterates.

\subsection{Homogeneous self-dual methods}

HSD embeddings provide a unified framework for recovering optimal solutions and certificates of infeasibility or unboundedness, as established by Ye
et al.~\cite{ye1994}. Skajaa et al.~\cite{skajaa2013}
developed warm-starting strategies for HSD interior-point
methods combining previously computed primal--dual solutions and a well-centered interior starting point. Modern homogeneous solvers include SCS~\cite{odonoghue2016},
which employs operator splitting, and
Clarabel~\cite{goulart2026}, which combines a homogeneous
embedding with a primal--dual interior-point method.
Although homogeneous formulations enable infeasibility
detection, their use within conventional IPMs does not
eliminate the interiority restrictions that complicate
warm-starting.

\subsection{Non-interior and proximal methods}

Nonlinear complementarity (NCP) functions provide an
alternative to explicitly enforcing complementarity
through logarithmic barriers. Proximal regularization
has also been employed to improve numerical robustness,
as in PIQP~\cite{schwan2023} and
QPDO~\cite{demarchi2022}. More closely related to our approach, ODYN~\cite{rojas2026}
derives all-shifted NCP functions from a proximal
penalty--barrier formulation for quadratic programming,
enabling non-interior iterates and effective warm-starting.
HALO-LP extends this interpretation to the homogeneous
complementarity pair by deriving the all-shifted HSD
centrality equations from a generalized penalty--barrier
relation. This preserves the ability to recover infeasibility certificates while admitting a condensed Newton formulation based on a positive-definite matrix.

\subsection{Contributions}

We introduce HALO-LP, a homogeneous all-shifted non-interior-point method for linear programming designed for warm-started optimization while retaining the optimality and infeasibility certification properties of homogeneous self-dual embeddings. Our main contributions are:

\begin{enumerate}
\renewcommand{\labelenumi}{(\roman{enumi})}

\item a homogeneous all-shifted formulation that combines
the HSD embedding with proximal primal--dual regularization
and shifted NCP functions, allowing primal--dual iterates to
leave the positive orthant and enabling arbitrary warm starts.
The resulting Newton equations admit a condensed formulation
requiring solves with a positive-definite matrix that can be
efficiently factorized using Cholesky decomposition;

\item a generalized penalty--barrier derivation of the all-shifted
centrality equations for both the standard complementarity pair
$(\vs,\vz)$ and the homogeneous pair $(\tau,\kappa)$;

\item an evaluation of HALO-LP on perturbed Netlib LP
instances, analytically characterized optimal and infeasible
problems, and sequential linear programming for centroidal
quadrupedal locomotion, demonstrating its warm-starting
capabilities, reliable status detection, and applicability to
robotics.

\end{enumerate}

\section{Background}
\label{sec:background}

This section summarizes the mathematical ingredients underlying
HALO-LP. We first introduce the primal--dual LP and its
homogeneous self-dual embedding, and then review nonlinear
complementarity representations and homogeneous primal--dual
centrality.

\subsection{Homogeneous self-dual embedding}

The HSD embedding reformulates the primal--dual optimality
conditions into a single homogeneous system that also encodes
certificates of primal and dual infeasibility. Consider the linear program
\begin{equation}
\label{eq:lp}
\begin{aligned}
    \min_{\vx,\vs}\quad
        & \vecC^{\transpose}\vx \\
    \mathrm{s.t.}\quad
        & \matA\vx=\vecB,\\
        & \matG\vx+\vs=\vecH,\\
        & \vs\geq\vzero,
\end{aligned}
\end{equation}
where
$\vx\in\R^n$ is the free primal variable,
$\vs\in\R_+^p$ is the slack variable,
$\vecC\in\R^n$,
$\vecB\in\R^m$,
$\vecH\in\R^p$,
$\matA\in\R^{m\times n}$, and
$\matG\in\R^{p\times n}$. Its dual is
\begin{equation}
\label{eq:dual-lp}
\begin{aligned}
    \max_{\vy,\vz}\quad
        & -\vecB^{\transpose}\vy
          -\vecH^{\transpose}\vz \\
    \mathrm{s.t.}\quad
        & \matA^{\transpose}\vy
          +\matG^{\transpose}\vz
          +\vecC=\vzero,\\
        & \vz\geq\vzero,
\end{aligned}
\end{equation}
where $\vy\in\R^m$ is the free equality multiplier and
$\vz\in\R_+^p$ is the inequality multiplier.
Introducing the nonnegative homogeneous variables
$\tau,\kappa\in\R_+$, the HSD embedding of \eqref{eq:lp}--\eqref{eq:dual-lp} is
\begin{subequations}
\label{eq:hsd}
\begin{align}
    \matA^{\transpose}\vy
    +\matG^{\transpose}\vz
    +\vecC\tau
    &=\vzero,
    \label{eq:hsd-d}\\
    \matA\vx-\vecB\tau
    &=\vzero,
    \label{eq:hsd-e}\\
    \matG\vx+\vs-\vecH\tau
    &=\vzero,
    \label{eq:hsd-i}\\
    -\vecC^{\transpose}\vx
    -\vecB^{\transpose}\vy
    -\vecH^{\transpose}\vz
    -\kappa
    &=0,
    \label{eq:hsd-g}\\
    \vs\circ\vz&=\vzero,
    \qquad
    \tau\kappa=0,
    \label{eq:hsd-c}\\
    (\vs,\vz,\tau,\kappa)&\geq0.
    \label{eq:hsd-pos}
\end{align}
\end{subequations}
Finally, we denote the affine HSD residuals by
\begin{equation}
\label{eq:hsd-res}
\begin{aligned}
    \vr_d
    &=
    \matA^{\transpose}\vy
    +\matG^{\transpose}\vz
    +\vecC\tau,
    \\
    \vr_e
    &=
    \matA\vx-\vecB\tau,
    \\
    \vr_i
    &=
    \matG\vx+\vs-\vecH\tau,
    \\
    r_g
    &=
    -\vecC^{\transpose}\vx
    -\vecB^{\transpose}\vy
    -\vecH^{\transpose}\vz
    -\kappa.
\end{aligned}
\end{equation}

\subsection{NCP functions and homogeneous centrality}
\label{sec:ncp-centrality}
The residuals in \eqref{eq:hsd-res} encode the affine part of the HSD system, while the remaining conditions in
\eqref{eq:hsd} enforce nonnegativity and complementarity.
These conditions can be represented through an NCP function satisfying
\begin{equation}
    \vncpFun{}(\vs,\vz)=\vzero
    \quad\Longleftrightarrow\quad
    \vs\geq\vzero,\; \vz\geq\vzero,\; \vs\circ\vz=\vzero.
\end{equation}
A smooth NCP representation is given by the
Fischer--Burmeister function
\begin{equation}
\label{eq:fb}
    \vncpFun{}(\vs,\vz;\mu)
    :=
    \vs + \vz - \sqrt{\vs^{\circ2} + \vz^{\circ2} + 2\mu},
    \qquad
    \mu>0,
\end{equation}
For the homogeneous pair $(\tau,\kappa)$, the corresponding
scalar function is denoted by $\chi(\tau,\kappa;\mu)$.
As $\mu\rightarrow0$, these equations recover the
complementarity conditions in \eqref{eq:hsd-c}.

In homogeneous primal--dual interior-point methods, centrality
can be characterized through the gradient of a logarithmic
barrier~\cite{DahlAndersen2022}. For the nonnegative orthant,
the standard barrier is
$\mathcal{B}(\vs,\tau)
:=-\sum_{j=1}^{p}\log\vs_j-\log\tau$.
Defining
$\vecZeta=(\vs^{\transpose},\tau)^{\transpose}$ and
$\vlam=(\vz^{\transpose},\kappa)^{\transpose}$,
the primal--dual barrier-gradient relation is
$\vlam=-\mu\nabla_{\vecZeta}\mathcal{B}$, which yields
\begin{equation}
\label{eq:standard-centrality}
    \vs\circ\vz=\mu\bone,
    \qquad
    \tau\kappa=\mu .
\end{equation}
Here, $\bone$ denotes the vector of ones of appropriate dimension.
These relations characterize the standard perturbed
complementarity conditions underlying homogeneous
primal--dual centrality. In Sec.~\ref{sec:barrier-centrality}, this construction is
generalized through a proximal penalty--barrier potential to
obtain the all-shifted centrality equations of HALO-LP.

\section{All-Shifted Homogeneous Non-Interior-Point LP}
\label{sec:method}

This section develops the mathematical formulation of HALO-LP.
We first introduce a proximal regularization of the HSD embedding and derive the all-shifted centrality equations from a generalized penalty--barrier construction. We then derive the condensed Newton system, establish its positive-definite reduced formulation, and characterize the optimality and infeasibility conditions used by the solver.
\subsection{Proximal HSD formulation}
\label{sec:prox-hsd}

Define $\vct{u} = (\vx^{\transpose},\vy^{\transpose}, \vz^{\transpose},\tau)^{\transpose}$, $\vv =(\vzero^{\transpose},\vzero^{\transpose}, \vs^{\transpose},\kappa)^{\transpose}$ and $\vr_H = (\vr_d^\transpose,-\vr_e^\transpose,-\vr_i^\transpose,r_g)^\transpose$. The linear HSD equations admit the skew-symmetric operator
\begin{equation}
\label{eq:M}
\mtr{M}
:=
\begin{bmatrix}
    0 & \matA^{\transpose} & \matG^{\transpose} & \vecC\\
    -\matA & 0 & 0 & \vecB\\
    -\matG & 0 & 0 & \vecH\\
    -\vecC^{\transpose} &
    -\vecB^{\transpose} &
    -\vecH^{\transpose} & 0
\end{bmatrix},
\qquad
\mtr{M}^{\transpose}=-\mtr{M},
\end{equation}
for which $\mtr{M}\vct{u}-\vv = \vr_H$. Let $\vct{u}_{E} =(\xE^{\transpose},\yE^{\transpose}, \zE^{\transpose},\tau_E)^{\transpose}$ and $\mtr{R} = \diag\left(\rhod\Id,\,\rhoe\Id,\,\rhoi\Id,\,\rho_\tau\right)\succ0$. We regularize the affine HSD system as
\begin{equation}
\label{eq:prox-hsd}
    \mtr{M}\vct{u}
    -\vv
    +\mtr{R}
    (\vct{u}-\vct{u}_{E})
    =
    \vzero .
\end{equation}
Since $\mtr{M}$ is skew-symmetric,
\begin{equation}
\label{eq:strong-monotone}
    \operatorname{sym}
    (\mtr{M}+\mtr{R})
    =
    \mtr{R}
    \succ0,
\end{equation}
so the proximal perturbation gives the affine HSD operator a
positive-definite symmetric part. Using the residual convention
in \eqref{eq:hsd-res}, the corresponding proximal residuals are
\begin{equation}
\label{eq:prox-res}
\begin{aligned}
    \tilde{\vr}_d
    &=
    \vr_d + \rho_d(\vx-\xE),
    &
    \tilde{\vr}_e
    &=
    \vr_e - \rho_e(\vy-\yE),
    \\
    \tilde{\vr}_i
    &=
    \vr_i - \rho_i(\vz-\zE),
    &
    \tilde{r}_g
    &=
    r_g + \rho_\tau(\tau-\tau_E).
\end{aligned}
\end{equation}

\subsection{Generalized penalty--barrier centrality}
\label{sec:barrier-centrality}

We next derive the all-shifted complementarity equations from
a generalized penalty--barrier construction. Following the
barrier-gradient interpretation of homogeneous primal--dual
centrality~\cite{DahlAndersen2022}, we use the augmented HSD
complementarity pair $(\vecZeta,\vlam)$ defined in
Sec.~\ref{sec:ncp-centrality} and introduce the proximal references
$\vecZetaE=(\sE^{\transpose},\tauE)^{\transpose}$ and
$\vlamE=(\zE^{\transpose},\kappaE)^{\transpose}$.
Let $\vxi\in\R_{++}^{p+1}$ denote an auxiliary barrier
variable. We then define the generalized penalty--barrier potential
\begin{equation}
\label{eq:penalty-barrier}
\begin{aligned}
\mathcal{P}_{\mu,\rhon}(\vecZeta,\vxi;\vlam)
={}&
-\mu\sum_{j=1}^{p+1}\log \xi_j
+\frac{1}{2}\norm{\vecZeta-\vxi}_2^2
-\vlam^{\transpose}(\vecZeta-\vxi)
\\
&+
\frac{\rhon}{4}
\norm{
(\vecZeta-\vecZetaE)+(\vlam-\vlamE)
}_2^2 .
\end{aligned}
\end{equation}
where $\rho_n$ denotes the NCP proximal regularization. To express the corresponding barrier-gradient relation in
normalized form, define
\begin{equation}
    \calF_{\mu,\rhon}
    :=
    \frac{1}{\mu}
    \calP_{\mu,\rhon}.
\end{equation}
We characterize homogeneous centrality through the generalized
barrier-gradient relations
\begin{equation}
\label{eq:generalized-centrality}
    \nabla_{\vxi}
        \calP_{\mu,\rhon}
    =
    \vzero,
    \qquad
    \vlam
    =
    -\mu
    \nabla_{\vecZeta}
        \calF_{\mu,\rhon}.
\end{equation}
The first condition in \eqref{eq:generalized-centrality}
gives
\begin{equation}
\label{eq:xi-quadratic}
    \vxi^{\circ 2}
    -
    (\vecZeta-\vlam)\circ\vxi
    -
    \mu\bone
    =
    \vzero,
\end{equation}
whose positive solution is
\begin{equation}
\label{eq:xi-positive}
    \vxi
    =
    \frac{1}{2}
    \left[
        \vecZeta-\vlam
        +
        \sqrt{
            (\vecZeta-\vlam)^{\circ 2}
            +4\mu\bone
        }
    \right].
\end{equation}
Here and throughout this subsection, vector powers and square
roots are evaluated componentwise. The second relation in \eqref{eq:generalized-centrality} yields
\begin{equation}
\label{eq:centrality-shift}
    \vecZeta-\vxi
    +\frac{\rhon}{2}
    \left[
        (\vecZeta-\vecZetaE)
        +(\vlam-\vlamE)
    \right]
    =\vzero .
\end{equation}
Eliminating $\vxi$ using \eqref{eq:xi-positive} gives
\begin{equation}
\label{eq:shifted-min}
    \vncpFun{}(\vecZeta,\vlam;\mu)
    +\rho_n(\vecZeta-\vecZetaE)
    +\rho_n(\vlam-\vlamE)
    =\vzero ,
\end{equation}
where the vector-valued smoothed minimum NCP function is
\begin{equation}
\label{eq:min-ncp}
\begin{aligned}
    \vncpFun{}
    \!\left(
        \vecZeta,\vlam;\mu
    \right)
    &=
    \vecZeta+\vlam
    -
    \sqrt{
        (\vecZeta-\vlam)^{\circ 2}
        +4\mu\bone
    }.
\end{aligned}
\end{equation}
Thus, the all-shifted relation follows directly from the
generalized penalty--barrier construction rather than being
introduced as an external perturbation.

Using the centrality relation
$\vecZeta\circ\vlam=\mu\bone$, the radicand in
\eqref{eq:min-ncp} satisfies
\[
    (\vecZeta-\vlam)^{\circ2}+4\mu\bone
    =
    \vecZeta^{\circ2}+\vlam^{\circ2}+2\mu\bone .
\]
Hence, the smoothed minimum NCP reduces to the smoothed
Fischer--Burmeister representation in \eqref{eq:fb}, as in
ODYN~\cite{rojas2026}. Splitting the augmented pair into
$(\vs,\vz)$ and $(\tau,\kappa)$, the all-shifted centrality
equations are
\begin{subequations}
\label{eq:all-shifted-ncp}
\begin{align}
    \tilde{\vr}_n
    &:=
    \vncpFun{}(\vs,\vz;\mu)
    +\rhon(\vs-\vs_E)
    +\rhon(\vz-\vz_E),
    \label{eq:all-shifted-ncp-sz}
    \\
    \tilde r_h
    &:=
    \chi(\tau,\kappa;\mu)
    +\rhon(\tau-\tau_E)
    +\rhon(\kappa-\kappa_E).
    \label{eq:all-shifted-ncp-tk}
\end{align}
\end{subequations}
Together, \eqref{eq:prox-res} and
\eqref{eq:all-shifted-ncp} define the proximal all-shifted
HSD equations solved by HALO-LP.
\subsection{Newton system and condensation}

For the Newton system, the equality and inequality block equations are multiplied by \(-1\), yielding an equivalent KKT-like form convenient for condensation. Therefore, the search direction is computed as:
\begin{equation}
\label{eq:newton-full}
\begin{bmatrix}
    \rhod\Id
        & \matA^{\transpose}
        & \matG^{\transpose}
        & 0
        & \vecC
        & 0
    \\
    \matA
        & -\rhoe\Id
        & 0
        & 0
        & -\vecB
        & 0
    \\
    \matG
        & 0
        & -\rhoi\Id
        & \Id
        & -\vecH
        & 0
    \\
    -\vecC^{\transpose}
        & -\vecB^{\transpose}
        & -\vecH^{\transpose}
        & 0
        & \rho_\tau
        & -1
    \\
    0
        & 0
        & \pertncpJac{z}
        & \pertncpJac{s}
        & 0
        & 0
    \\
    0
        & 0
        & 0
        & 0
        & \widetilde{\chi}_{\tau}
        & \widetilde{\chi}_{\kappa}
\end{bmatrix}
\begin{bmatrix}
    \dx\\
    \dy\\
    \dz\\
    \ds\\
    \Delta\tau\\
    \Delta\kappa
\end{bmatrix}
=
-
\begin{bmatrix}
    \tilde{\vr}_d\\
    \tilde{\vr}_e\\
    \tilde{\vr}_i\\
    \tilde{r}_g\\
    \tilde{\vr}_n\\
    \tilde{r}_h
\end{bmatrix}.
\end{equation}
where $\pertncpJac{s} = \ncpJac{s}+\rhon\Id$, $\pertncpJac{z} = \ncpJac{z}+\rhon\Id$, $\widetilde{\chi}_{\tau} = \chi_{\tau}+\rhon$ and $\widetilde{\chi}_{\kappa} = \chi_{\kappa}+\rhon$ denote regularized NCP Jacobians. The complementary directions are eliminated as
\begin{equation}
\label{eq:comp-elim}
\begin{aligned}
    \ds
    &=
    -\pertncpJac{s}^{-1}
    \left(
        \tilde{\vr}_n
        +\pertncpJac{z}\dz
    \right),
    \\
    \Delta\kappa
    &=
    -\widetilde{\chi}_{\kappa}^{-1}
    \left(
        \tres{h}
        +\widetilde{\chi}_{\tau}\Delta\tau
    \right).
\end{aligned}
\end{equation}

Substitution of \eqref{eq:comp-elim} into
\eqref{eq:newton-full} yields
\begin{equation}
\label{eq:newton-condensed}
\begin{bmatrix}
    \rhod\Id
        & \matA^{\transpose}
        & \matG^{\transpose}
        & \vecC
    \\
    \matA
        & -\rhoe\Id
        & 0
        & -\vecB
    \\
    \matG
        & 0
        & -\matD
        & -\vecH
    \\
    -\vecC^{\transpose}
        & -\vecB^{\transpose}
        & -\vecH^{\transpose}
        & \gamma_\tau
\end{bmatrix}
\begin{bmatrix}
    \dx\\
    \dy\\
    \dz\\
    \Delta\tau
\end{bmatrix}
= -
\begin{bmatrix}
    \tilde{\vr}_d\\
    \tilde{\vr}_e\\
    \bar{\vr}_i\\
    \bar{r}_g
\end{bmatrix}.
\end{equation}
where $\bar{\vr}_i = \tilde{\vr}_i - \pertncpJac{s}^{-1}\tilde{\vr}_n$ is the condensed right-hand side used to compute $\dz$, $\bar{r}_g = \tilde{r}_g + \widetilde{\chi}_{\kappa}^{-1}\tres{h}$ is the condensed right-hand side used to compute $\Delta\tau$, $\matP = \pertncpJac{s}^{-1}\pertncpJac{z}$, $\matD = \matP + \rhoi\Id$ and $\gamma_\tau = \rho_\tau + \tilde{\chi}_{\kappa}^{-1}\tilde{\chi}_{\tau}$. 

\subsection{Positive-definite reduced solve}

The equality and inequality dual directions obtained from
\eqref{eq:newton-condensed} are
\begin{equation}
\label{eq:dual-elim}
\begin{aligned}
    \dy
    &=
    \frac{1}{\rhoe}
    \left(
        \matA\dx
        -\vecB\Delta\tau
        +\tilde{\vr}_e
    \right),
    \\
    \dz
    &=
    \matD^{-1}
    \left(
        \matG\dx
        -\vecH\Delta\tau
        +\bar{\vr}_i
    \right).
\end{aligned}
\end{equation}
Define
\begin{equation}
\label{eq:H-condensed}
    \matH
    =
    \rhod\Id
    +
    \frac{1}{\rhoe}
    \matA^{\transpose}\matA
    +
    \matG^{\transpose}
    \matD^{-1}
    \matG,
\end{equation}
together with
\begin{equation}
\label{eq:homogeneous-coupling}
\begin{aligned}
    \vct{a}
    &=
    \vecC
    -
    \frac{1}{\rhoe}
    \matA^{\transpose}\vecB
    -
    \matG^{\transpose}\matD^{-1}\vecH,
    \\
    \vct{d}
    &=
    -\vecC
    -
    \frac{1}{\rhoe}
    \matA^{\transpose}\vecB
    -
    \matG^{\transpose}\matD^{-1}\vecH,
    \\
    \gamma
    &=
    \gamma_\tau
    +
    \frac{1}{\rhoe}
    \vecB^{\transpose}\vecB
    +
    \vecH^{\transpose}\matD^{-1}\vecH.
\end{aligned}
\end{equation}
The corresponding right-hand sides are
\begin{equation}
\label{eq:reduced-rhs}
\begin{aligned}
    \bar{\vr}_x
    &=
    \tilde{\vr}_d
    +
    \frac{1}{\rhoe}
    \matA^{\transpose}\tilde{\vr}_e
    +
    \matG^{\transpose}\matD^{-1}\bar{\vr}_i,
    \\
    \bar{r}_\tau
    &=
    \bar{r}_g
    -
    \frac{1}{\rhoe}
    \vecB^{\transpose}\tilde{\vr}_e
    -
    \vecH^{\transpose}\matD^{-1}\bar{\vr}_i.
\end{aligned}
\end{equation}
Therefore, the Newton system reduces to
\begin{equation}
\label{eq:reduced-hsd}
    \begin{bmatrix}
        \matH & \vct{a}\\
        \vct{d}^{\transpose} & \gamma
    \end{bmatrix}
    \begin{bmatrix}
        \dx\\
        \Delta\tau
    \end{bmatrix}
    =-
    \begin{bmatrix}
        \bar{\vr}_x\\
        \bar{r}_\tau
    \end{bmatrix}.
\end{equation}

The scalar homogeneous direction can be separated using two
linear systems with the same matrix $\matH$:
\begin{equation}
\label{eq:two-solves}
    \matH\vp = -\vct{a},
    \qquad
    \matH\vq = -\bar{\vr}_x.
\end{equation}
It follows that
\begin{equation}
\label{eq:dtau}
    \Delta\tau
    =
    \frac{
        -\bar{r}_\tau
        -\vct{d}^{\transpose}\vq
    }{
        \gamma
        +\vct{d}^{\transpose}\vp
    },
\end{equation}
and
\begin{equation}
\label{eq:dx-recover}
    \dx
    =
    \vq
    +\Delta\tau\,\vp.
\end{equation}
The remaining directions are recovered from
\eqref{eq:dual-elim} and \eqref{eq:comp-elim}.

For the smoothed NCP functions considered here, the diagonal
Jacobians can be written as
\begin{equation}
\label{eq:ncp-jacobians}
    \ncpJac{s}
    =
    \Id-\mtr{R}_n^{-1}\mtr{C}_{s},
    \qquad
    \ncpJac{z}
    =
    \Id-\mtr{R}_n^{-1}\mtr{C}_{z},
\end{equation}
where $\mtr{R}_n\succ0$, $\mtr{C}_{s}$, and $\mtr{C}_{z}$
are diagonal, and the smoothing term $\mu>0$ ensures
\[
    -\Id
    \prec
    \mtr{R}_n^{-1}\mtr{C}_{s},
    \mtr{R}_n^{-1}\mtr{C}_{z}
    \prec
    \Id .
\]
Therefore,
\begin{equation}
\label{eq:ncp-jacobian-bound}
    \vzero
    \prec
    \ncpJac{s},\ncpJac{z}
    \prec
    2\Id .
\end{equation}
Since
$\pertncpJac{s}=\ncpJac{s}+\rhon\Id$ and
$\pertncpJac{z}=\ncpJac{z}+\rhon\Id$
are diagonal, \eqref{eq:ncp-jacobian-bound} implies
\begin{equation}
\label{eq:P-bound}
    \matP
    =
    \pertncpJac{s}^{-1}\pertncpJac{z}
    \succeq
    \frac{\rhon}{\rhon+2}\Id .
\end{equation}
Consequently,
\begin{equation}
\label{eq:D-positive}
    \matD
    =
    \matP+\rhoi\Id
    \succeq
    \left(
        \rhoi+\frac{\rhon}{\rhon+2}
    \right)\Id
    \succ0 .
\end{equation}
Hence, for any $\vv\neq\vzero$,
\begin{equation}
\label{eq:H-positive}
    \vv^{\transpose}\matH\vv
    =
    \rhod\norm{\vv}_2^2
    +\frac{1}{\rhoe}\norm{\matA\vv}_2^2
    +(\matG\vv)^{\transpose}
      \matD^{-1}(\matG\vv)
    >0 ,
\end{equation}
which establishes $\matH\succ0$.
Thus, $\matH$ admits a Cholesky factorization
$\matH=\matL\matL^{\transpose}$, which is reused for both
right-hand sides in \eqref{eq:two-solves}.

At a proximal fixed point,
\begin{equation}
    \vx=\xE,\qquad
    \vy=\yE,\qquad
    \vz=\zE,\qquad
    \vs=\sE,
\end{equation}
together with $\tau=\tau_E$ and $\kappa=\kappa_E$.
Hence, as $\mu\to0$, the proximal and all-shifted
perturbations in \eqref{eq:prox-res} and
\eqref{eq:all-shifted-ncp} vanish, recovering the original HSD
conditions \eqref{eq:hsd}. Moreover, setting the proximal terms to zero
in the full Newton system \eqref{eq:newton-full} recovers the
non-proximal HSD Newton equations. The positive-definite
condensation \eqref{eq:H-condensed}, however, explicitly
exploits the proximal regularization.

\subsection{Optimality and infeasibility certificates}
\label{sec:hsd-certificates}

The HSD embedding \eqref{eq:hsd} represents both primal--dual
solutions and infeasibility certificates within the same system.
When $\tau>0$, complementarity implies $\kappa=0$, and the
scaled variables
\begin{equation}
\label{eq:hsd-recovery}
    \hat{\vx}:=\frac{\vx}{\tau},
    \qquad
    \hat{\vy}:=\frac{\vy}{\tau},
    \qquad
    \hat{\vz}:=\frac{\vz}{\tau},
    \qquad
    \hat{\vs}:=\frac{\vs}{\tau}
\end{equation}
satisfy the primal--dual KKT conditions of
\eqref{eq:lp}--\eqref{eq:dual-lp}.

Conversely, when $\tau=0$ and $\kappa>0$, the HSD equations
reduce to
\begin{subequations}
\label{eq:hsd-cert}
\begin{align}
    \matA^{\transpose}\vy+\matG^{\transpose}\vz
    &=\vzero,
    \qquad
    \vz\geq\vzero,
    \label{eq:cert-primal}\\
    \matA\vx&=\vzero,
    \label{eq:cert-dual-a}\\
    \matG\vx+\vs&=\vzero,
    \qquad
    \vs\geq\vzero,
    \label{eq:cert-dual-b}\\
    -\vecC^{\transpose}\vx
    -\vecB^{\transpose}\vy
    -\vecH^{\transpose}\vz
    &=\kappa>0.
    \label{eq:cert-gap}
\end{align}
\end{subequations}
Equation \eqref{eq:cert-gap} can be written as
\begin{equation}
\label{eq:cert-split}
    \kappa
    =
    -\vecC^{\transpose}\vx
    -
    \left(
        \vecB^{\transpose}\vy
        +\vecH^{\transpose}\vz
    \right).
\end{equation}
Hence, at least one of the two terms on the right-hand side is
strictly positive.

If $\vecB^{\transpose}\vy+\vecH^{\transpose}\vz<0$, then
\begin{equation}
\label{eq:primal-farkas}
    \matA^{\transpose}\vy+\matG^{\transpose}\vz=\vzero,
    \qquad
    \vz\geq\vzero,
    \qquad
    \vecB^{\transpose}\vy+\vecH^{\transpose}\vz<0
\end{equation}
is a Farkas certificate of primal infeasibility. Indeed, for any
primal-feasible point $\vx_f$,
\begin{equation}
    0
    =
    \vx_f^{\transpose}
    \left(
        \matA^{\transpose}\vy
        +\matG^{\transpose}\vz
    \right)
    \leq
    \vecB^{\transpose}\vy
    +\vecH^{\transpose}\vz,
\end{equation}
contradicting $\vecB^{\transpose}\vy+\vecH^{\transpose}\vz<0$.

Similarly, if $\vecC^{\transpose}\vx<0$, then
\eqref{eq:hsd-cert} implies
\begin{equation}
\label{eq:dual-farkas}
    \matA\vx=\vzero,
    \qquad
    \matG\vx\leq\vzero,
    \qquad
    \vecC^{\transpose}\vx<0,
\end{equation}
which is a certificate of dual infeasibility. If the primal problem
is feasible, $\vx$ is also a recession direction along which the
objective decreases without bound.

Therefore, the HSD variables distinguish between optimality and
infeasibility through
\begin{equation}
\label{eq:hsd-cases}
\begin{aligned}
    \tau>0
    &\quad\Longrightarrow\quad
    \text{primal--dual solution},\\
    \tau=0,\ \kappa>0,\ 
    \vecB^{\transpose}\vy+\vecH^{\transpose}\vz<0
    &\quad\Longrightarrow\quad
    \text{primal infeasibility},\\
    \tau=0,\ \kappa>0,\ 
    \vecC^{\transpose}\vx<0
    &\quad\Longrightarrow\quad
    \text{dual infeasibility}.
\end{aligned}
\end{equation}

\subsection{Stopping criteria}
\label{sec:stopping_criteria}

For $\tau>0$, a primal--dual solution of the original LP is
recovered by scaling the homogeneous variables by $\tau$.
Rather than explicitly forming these scaled variables, we test
equivalent termination conditions directly in the homogeneous
coordinates, thereby avoiding division by potentially small values
of $\tau$.

The algorithm declares convergence when
\begin{equation}
\label{eq:optimality-stop}
\begin{aligned}
\norm{\vr_d}_{\infty}
&\leq
\epsa\tau
+
\epsr
\max\!
\left(
    \tau\norm{\vecC}_{\infty},
    \norm{\matA^{\transpose}\vy}_{\infty},
    \norm{\matG^{\transpose}\vz}_{\infty}
\right),
\\
\norm{\vr_e}_{\infty}
&\leq
\epsa\tau
+
\epsr
\max\!
\left(
    \norm{\matA\vx}_{\infty},
    \tau\norm{\vecB}_{\infty}
\right),
\\
\norm{\vr_i}_{\infty}
&\leq
\epsa\tau
+
\epsr
\max\!
\left(
    \norm{\matG\vx}_{\infty},
    \norm{\vs}_{\infty},
    \tau\norm{\vecH}_{\infty}
\right),
\\
\norm{\vr_c}_{\infty}
&\leq
\epsa\tau
+
\epsr
\max\!
\left(
    \norm{\vs}_{\infty},
    \norm{\vz}_{\infty},
    \norm{\vr_n}_{\infty}
\right),
\\
\abs{r_g}
&\leq
\epsa\tau
+
\epsr
\max\!
\left(
    \tau,
    \abs{\vecC^{\transpose}\vx},
    \abs{\vecB^{\transpose}\vy+\vecH^{\transpose}\vz}
\right),
\\
\abs{\kappa}
&\leq
(\epsa+\epsr)\tau .
\end{aligned}
\end{equation}
where $\vr_c = \min(\vs,\vz)$ denotes the exact complementarity
residual and $\vr_n = \vncpFun{}(\vs,\vz;\mu)$ denotes the NCP residual used for relative scaling. Here, $\epsa$ and $\epsr$ denote the absolute and relative termination tolerances, respectively, and $\min(\vs,\vz)$ is
evaluated componentwise. These conditions are equivalent to
applying the corresponding normalized criteria to
$(\vx,\vy,\vz,\vs)/\tau$, but avoid explicit division by $\tau$.

When $\tau$ is sufficiently small, the optimality test
\eqref{eq:optimality-stop} is not applied; instead, the HSD
infeasibility certificates described in
Sec.~\ref{sec:hsd-certificates} are evaluated.

\subsection{Infeasibility stopping criteria}
\label{sec:infeasibility-stopping}

Infeasibility certificates are tested when $\kappa>\epsa$ and $\abs{\tau}\leq(\epsa+\epsr)\kappa$. We use separate scale-relative tolerances $\epsp>0$ and $\epsd>0$ for primal and dual infeasibility, respectively.

For primal infeasibility, let
$\omega_p:=\max(\norm{\vy}_{\infty},\norm{\vz}_{\infty})$.
The certificate is accepted when
\begin{equation}
\label{eq:primal-infeas-stop}
\begin{gathered}
    \norm{\matA^{\transpose}\vy+\matG^{\transpose}\vz}_{\infty}
    \leq \epsp\omega_p,
    \qquad
    \norm{\min(\vz,\vzero)}_{\infty}
    \leq \epsp\omega_p,
    \\
    \vecB^{\transpose}\vy+\vecH^{\transpose}\vz
    < -\epsp\omega_p .
\end{gathered}
\end{equation}
For dual infeasibility, let
$\omega_d:=\norm{\vx}_{\infty}$.
The certificate is accepted when
\begin{equation}
\label{eq:dual-infeas-stop}
\begin{array}{c@{\qquad}c}
    \displaystyle
    \norm{\matA\vx}_{\infty}
    \leq \epsd\omega_d,
    &
    \displaystyle
    \vecC^{\transpose}\vx
    < -\epsd\omega_d,
    \\[1mm]
    \displaystyle
    \norm{\matG\vx+\vs}_{\infty}
    \leq \epsd\omega_d,
    &
    \displaystyle
    \norm{\min(\vs,\vzero)}_{\infty}
    \leq \epsd\omega_d .
\end{array}
\end{equation}
Conditions associated with absent equality or inequality
constraints are omitted.

\section{HALO-LP algorithm}

The Newton direction is globalized using the flexible
non-monotone Armijo line search introduced in
ODYN~\cite{rojas2026}. For HALO-LP, we employ the natural NCP merit function associated with the nonlinear complementarity
residuals,
\begin{equation}
\label{eq:natural-merit}
    \NatMerit(\vct{w})
    =
    \frac{1}{2}
    \left(
        \norm{\tres{n}}_{2}^{2}
        +
        \abs{\tres{h}}^{2}
    \right),
\end{equation}
where $\vw=(\vx^{\transpose},\vy^{\transpose},\vz^{\transpose},
\vs^{\transpose},\tau,\kappa)^{\transpose}$. The smoothing parameter is computed from the joint
complementarity violation of the augmented HSD pair
$(\vecZeta,\vlam)$ as
\begin{equation}
\label{eq:smoothing-parameter}
    \mu
    =
    \frac{
        \norm{\min(\vecZeta,\vlam)}_2^2
    }{
        p+1
    }.
\end{equation}
Collect the proximal regularization parameters in
$\boldsymbol{\rho}=(\rho_d,\rho_e,\rho_i,\rho_\tau,\rho_n)$
and the proximal primal--dual references in
$\vw_E=(\vx_E^{\transpose},\vy_E^{\transpose},
\vz_E^{\transpose},\vs_E^{\transpose},\tau_E,\kappa_E)^{\transpose}$.
The stacked perturbed HSD residual is
$\tilde{\vr}(\vw;\boldsymbol{\Pi})
=(\tilde{\vr}_d^{\transpose},
\tilde{\vr}_e^{\transpose},
\tilde{\vr}_i^{\transpose},
\tilde r_g,
\tilde{\vr}n^{\transpose},
\tilde r_h)^{\transpose}$,
where
$\boldsymbol{\Pi}:=(\mu,\boldsymbol{\rho},\vw_E)$.
Its infinity norm is denoted by
$\tilde{\mathcal{E}}(\vw;\boldsymbol{\Pi})
=\norm{\tilde{\vr}(\vw;\boldsymbol{\Pi})}_{\infty}$,
with $\tilde{\mathcal{E}}^{\mathrm{ref}}$ the corresponding
stored reference value. For a trial step length $\alpha$, let
$\vw^{+\alpha}:=\vw+\alpha\Delta\vw$ denote the trial iterate. Following ODYN~\cite{rojas2026}, an accepted
iterate triggers the parameter updates when
\begin{equation}
\label{eq:update-neighborhood}
    \tilde{\mathcal{E}}
    (\vw^{+\alpha};\boldsymbol{\Pi})
    \leq
    \theta\tilde{\mathcal{E}}^{\mathrm{ref}}
    +\beta\mu,
\end{equation}
where $\theta\in(0,1)$ specifies the required relative residual
reduction and $\beta>0$ weights the smoothing contribution.
The line search and the update rules for the smoothing parameter, step-length estimates, and proximal primal--dual references follow ODYN~\cite{rojas2026}, with the complementarity measure replaced by the homogeneous measure in \eqref{eq:smoothing-parameter}.

The complete HALO-LP solver is summarized in
Algorithm~\ref{alg:proposed-solver}.

\begin{algorithm}[H]
\caption{HALO-LP solver.}
\label{alg:proposed-solver}
\begin{algorithmic}[1]

\FOR{$k \gets 0$ \TO $k_{\max}-1$}

    \STATE search direction: $\Delta w_k$
    \hfill Eq.~\eqref{eq:newton-full}

    \STATE flexible Armijo line search: $\alpha_k$
    \hfill

    \IF{$\tilde{\mathcal{E}}(\vw^{+\alpha};\boldsymbol{\Pi}) 
            \leq \theta \tilde{\mathcal{E}}^{ref} + \beta\mu$}
        \STATE update smoothing parameter $\mu_{k+1}$
        \STATE update step-length estimates\\
        \STATE update proximal primal--dual estimates\\
        \STATE $\tilde{\mathcal{E}}^\mathrm{ref} \gets \tilde{\mathcal{E}}(\vw^{+\alpha};\boldsymbol{\Pi})$
    \ENDIF
    \STATE test optimality and infeasibility \hfill Eqs.~\eqref{eq:optimality-stop}--\eqref{eq:dual-infeas-stop}
\ENDFOR

\end{algorithmic}
\end{algorithm}

\section{Results}
We evaluate HALO-LP in three settings. First, we assess
its warm-starting performance against Clarabel, ProxQP,
and FBstab on perturbed LP instances derived from the
Netlib test set. Second, we evaluate its
ability to detect optimality, primal infeasibility,
and dual infeasibility using three analytically
characterized LP problems. Finally, we demonstrate
its applicability to robotic optimal control through
centroidal walking and trotting problems solved using SLP.

\subsection{Netlib Warm-Starting Benchmark}

We evaluate the warm-starting performance of HALO-LP
against Clarabel, ProxQP and FBstab on LP instances derived from the
Netlib test set~\cite{gay1985electronic}. Following the methodology
of Benson and Shanno~\cite{benson2007}, we first solve
each original LP from a cold start and reuse its
primal--dual solution to initialize perturbed instances.
The perturbations follow Engau et al.~\cite{engau2009},
with relative magnitudes
$\delta\in\{10^{-3},10^{-2},10^{-1}\}$, while preserving
the original sparsity structure.

For each perturbation level, we generate ten instances
and measure the warm-to-cold iteration ratio $\mathrm{WCR}_{p,s} = N^{\mathrm{warm}}_{p,s}/N^{\mathrm{cold}}_{p,s}$ where $N^{\mathrm{warm}}_{p,s}$ and $N^{\mathrm{cold}}_{p,s}$ denote the iterations required by solver $s$ on problem $p$ using warm and cold starts, respectively. Values below one indicate a reduction in iteration count due to warm-starting. Following Dolan and Mor\'e~\cite{dolan2002}, we define the performance ratio
\begin{equation}
\label{eq:wcr-performance-ratio}
    r_{p,s}
    =
    \frac{
        \mathrm{WCR}_{p,s}
    }{
        \displaystyle
        \min_{s'\in\mathcal{S}}
        \mathrm{WCR}_{p,s'}
    },
\end{equation}
and the corresponding performance profile
\begin{equation}
\label{eq:wcr-performance-profile}
    \rho_s(\tau)
    =
    \frac{1}{|\mathcal{P}|}
    \left|
        \left\{
            p\in\mathcal{P}
            :
            r_{p,s}\leq\tau
        \right\}
    \right|.
\end{equation}

\begin{figure*}[!t]
    \centering
    \begin{minipage}[t]{0.31\textwidth}
        \centering
        \subfloat[$\delta = 0.001$]{
            \includegraphics[width=\linewidth]{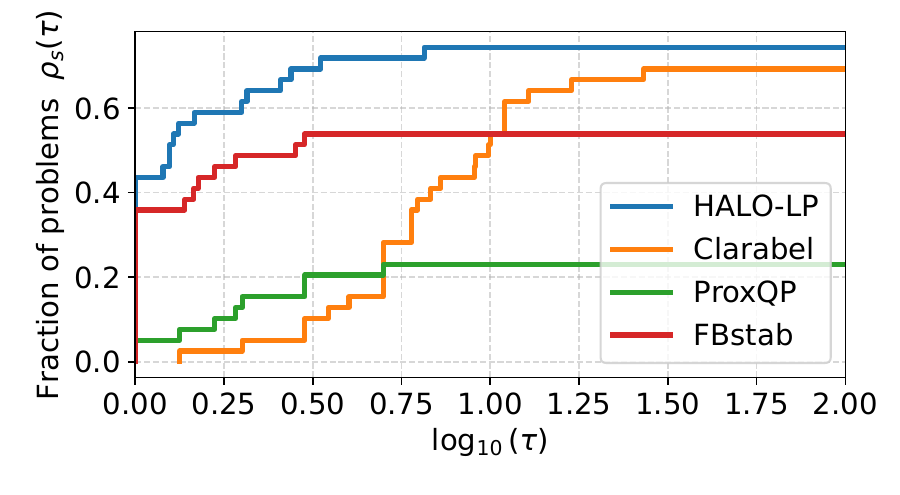}
            \label{fig:mm_warm_delta_0001}
        }
    \end{minipage}%
    \hfill%
    \begin{minipage}[t]{0.31\textwidth}
        \centering
        \subfloat[$\delta = 0.01$]{
            \includegraphics[width=\linewidth]{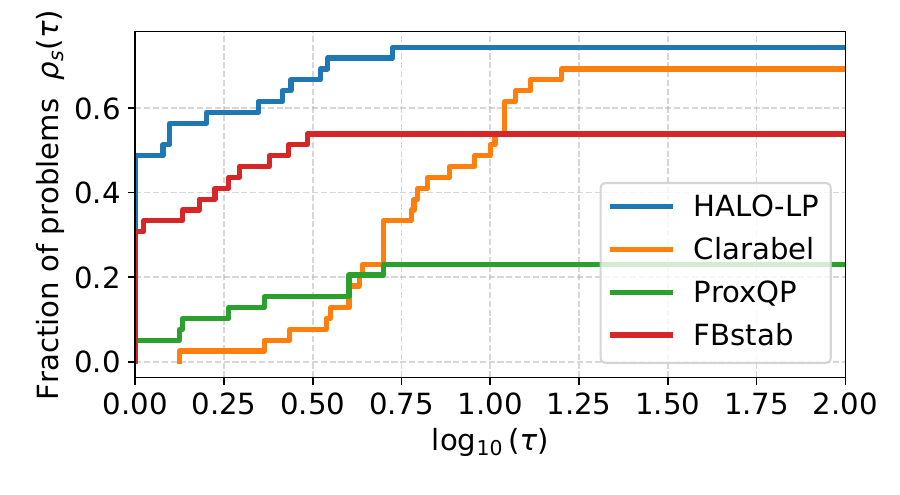}
            \label{fig:mm_warm_delta_001}
        }
    \end{minipage}%
    \hfill%
    \begin{minipage}[t]{0.31\textwidth}
        \centering
        \subfloat[$\delta = 0.1$]{
            \includegraphics[width=\linewidth]{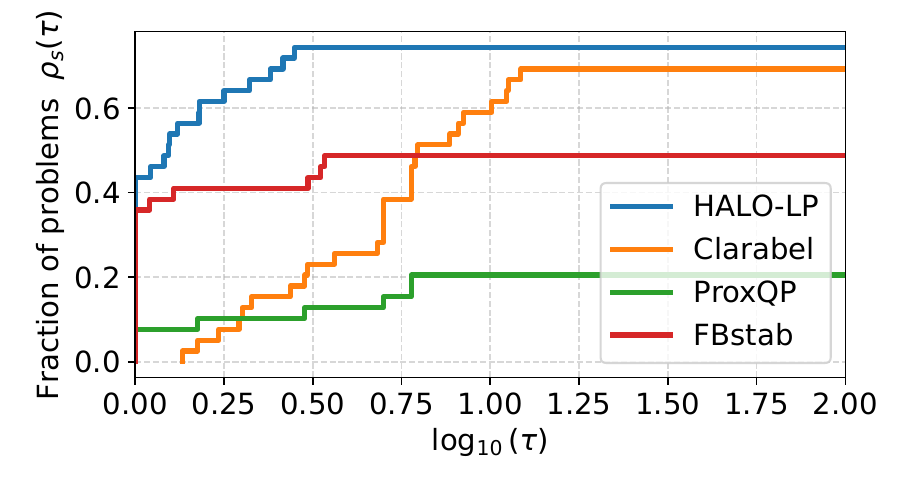}
            \label{fig:mm_warm_delta_01}
        }
    \end{minipage}%
    \caption{Warm-starting performance on perturbed instances
    of the Netlib test set for different
    perturbation magnitudes:
    (a) $\delta = 0.001$,
    (b) $\delta = 0.01$, and
    (c) $\delta = 0.1$.}
    \label{fig:mm_warmstarting}
\end{figure*}

Fig.~\ref{fig:mm_warmstarting} reports performance
profiles based on the WCR. HALO-LP achieves the best
warm-starting ratio on the largest fraction of tested
instances across all perturbation levels, demonstrating
its ability to exploit solutions of closely related LPs.

\subsection{Optimality and Infeasibility Detection}

To evaluate HALO-LP's ability to distinguish optimality
from infeasibility, we consider three LP problems with
analytically known solutions or certificates:
\begin{subequations}\label{eq:toy_lps}
\begin{align}
\mathrm{(P1)}\quad&
\begin{aligned}[t]
\min_{x_1,x_2}\quad &x_1+2x_2\\
\mathrm{s.t.}\quad &x_1+x_2\ge 1,\\
&0\le x_1,x_2\le 2,
\end{aligned}
\label{eq:toy_opt}\\
\mathrm{(P2)}\quad&
\begin{aligned}[t]
\min_{x_1,x_2}\quad &x_1\\
\mathrm{s.t.}\quad &x_1\le 0,\quad x_1\ge 1,\\
&0\le x_2\le 1,
\end{aligned}
\label{eq:toy_pinf}\\
\mathrm{(P3)}\quad&
\begin{aligned}[t]
\min_{x_1,x_2}\quad &-x_1\\
\mathrm{s.t.}\quad &x_1\ge 0,\quad 0\le x_2\le 1.
\end{aligned}
\label{eq:toy_dinf}
\end{align}
\end{subequations}
Problem~\eqref{eq:toy_opt} has the optimal solution
$(x_1,x_2)=(1,0)$ with objective value $1$.
Problem~\eqref{eq:toy_pinf} is primal infeasible
due to contradictory bounds, whereas
Problem~\eqref{eq:toy_dinf} is primal unbounded
along the feasible direction $(1,0)$ and thus
dual infeasible. Table~\ref{tab:toy_lp} summarizes the classification results and solver-reported iterations for the three test problems.

\begin{table}[!t]
\centering
\caption{Status agreement and solver-reported iterations
on characterized LP problems. A check mark indicates correct
classification.}
\label{tab:toy_lp}
{\scriptsize
\setlength{\tabcolsep}{2.5pt}
\renewcommand{\arraystretch}{1.15}
\begin{tabular}{lccccc}
\hline
Case & HALO-LP & ProxQP & Clarabel & FBstab & Gurobi\\
\hline
Optimal
& $\checkmark\,5$ & $\checkmark\,14$
& $\checkmark\,5$ & $\checkmark\,5$
& $\checkmark\,5$\\
Primal infeas.
& $\checkmark\,4$ & $\checkmark\,17$
& $\checkmark\,6$ & $\checkmark\,7$
& $\checkmark\,3$\\
Dual infeas.
& $\checkmark\,4$ & $\times\,692$
& $\checkmark\,6$ & $\checkmark\,6$
& $\checkmark\,4$\\
\hline
\end{tabular}
}
\end{table}

All problems are solved with a common tolerance of
$10^{-7}$ and an iteration cap of $500$. Presolve is disabled for Clarabel and Gurobi.
HALO-LP correctly classifies all three problems within
$4$--$5$ iterations, while Clarabel, FBstab, and Gurobi also
identify all problem statuses. ProxQP reaches its
iteration limit on the dual-infeasible case, with
$692$ cumulative inner iterations. Since iteration
counters differ across solvers, with FBstab reporting
cumulative semismooth Newton iterations, these results
primarily assess status detection.

\subsection{HALO-LP for optimal control}

We evaluate HALO-LP within an SLP framework for centroidal locomotion. The state
is
$\vx_k=(\vct{c}_k^{\transpose},
\boldsymbol{\ell}_k^{\transpose},
\vct{h}_k^{\transpose})^{\transpose}$,
where $\vct{c}_k$, $\boldsymbol{\ell}_k$, and $\vct{h}_k$ denote
the center of mass (CoM), linear momentum, and angular momentum,
respectively. The control
$\control_k=(\vct{f}_{1,k}^{\transpose},\ldots,
\vct{f}_{4,k}^{\transpose})^{\transpose}$
collects the contact forces. The centroidal dynamics are
\begin{equation}
\label{eq:centroidal-dynamics}
    \dot{\vct{c}}
    =
    \frac{\boldsymbol{\ell}}{m},
    \qquad
    \dot{\boldsymbol{\ell}}
    =
    \sum_i \vct{f}_i + m\vct{g},
    \qquad
    \dot{\vct{h}}
    =
    \sum_i (\vp_i-\vct{c})\times\vct{f}_i,
\end{equation}
where $m$ is the robot mass, $\vct{g}$ is the gravity vector, and
$\vp_i$ is the prescribed position of contact $i$.

We use a quadratic state- and force-tracking objective
subject to discretized centroidal dynamics, friction pyramids,
force bounds, and prescribed contact schedules. At each SLP iteration, first-order Taylor models of the objective
and constraints yield the elastic LP
\begin{equation}
\label{eq:slp-lp}
\begin{aligned}
    \min_{\vct{\delta},\vct{e}^+,\vct{e}^-,\vv}\quad
        & \widetilde{\vecC}^{\transpose}\vct{\delta}
        +\rho_{\mathrm{el}}\bone^{\transpose}
        (\vct{e}^+ + \vct{e}^- + \vv) \\
    \mathrm{s.t.}\quad
        & \widetilde{\matA}\bar{\matD}\vct{\delta}
        -\vct{e}^+ + \vct{e}^- = \widetilde{\vecB},\\
        & \widetilde{\matG}\bar{\matD}\vct{\delta}
        -\vv \leq \widetilde{\vecH},\\
        & \norm{\vct{\delta}}_{\infty}\leq\Delta,\\
        & \vct{e}^+,\vct{e}^-,\vv\geq\vzero .
\end{aligned}
\end{equation}
Here, $\vct{\delta}$ is the scaled SLP step,
$\widetilde{\vecC}$ is the linearized objective vector,
$\widetilde{\matA}$, $\widetilde{\vecB}$,
$\widetilde{\matG}$, and $\widetilde{\vecH}$ denote the
linearized constraint data, and $\vct{e}^+$, $\vct{e}^-$, and $\vv$
are elastic feasibility variables. The matrix $\bar{\matD}$ scales
the optimization variables, $\Delta$ is the trust-region radius,
and $\rho_{\mathrm{el}}=10$ penalizes constraint violations.
Each SLP subproblem is supplied directly to HALO-LP,
while a merit-function line search globalizes the nonlinear
SLP iterations.

\begin{figure}[!t]
    \centering
    \subfloat[Walking motion]{
        \begin{minipage}{0.98\columnwidth}
            \centering
            \includegraphics[width=0.32\linewidth]{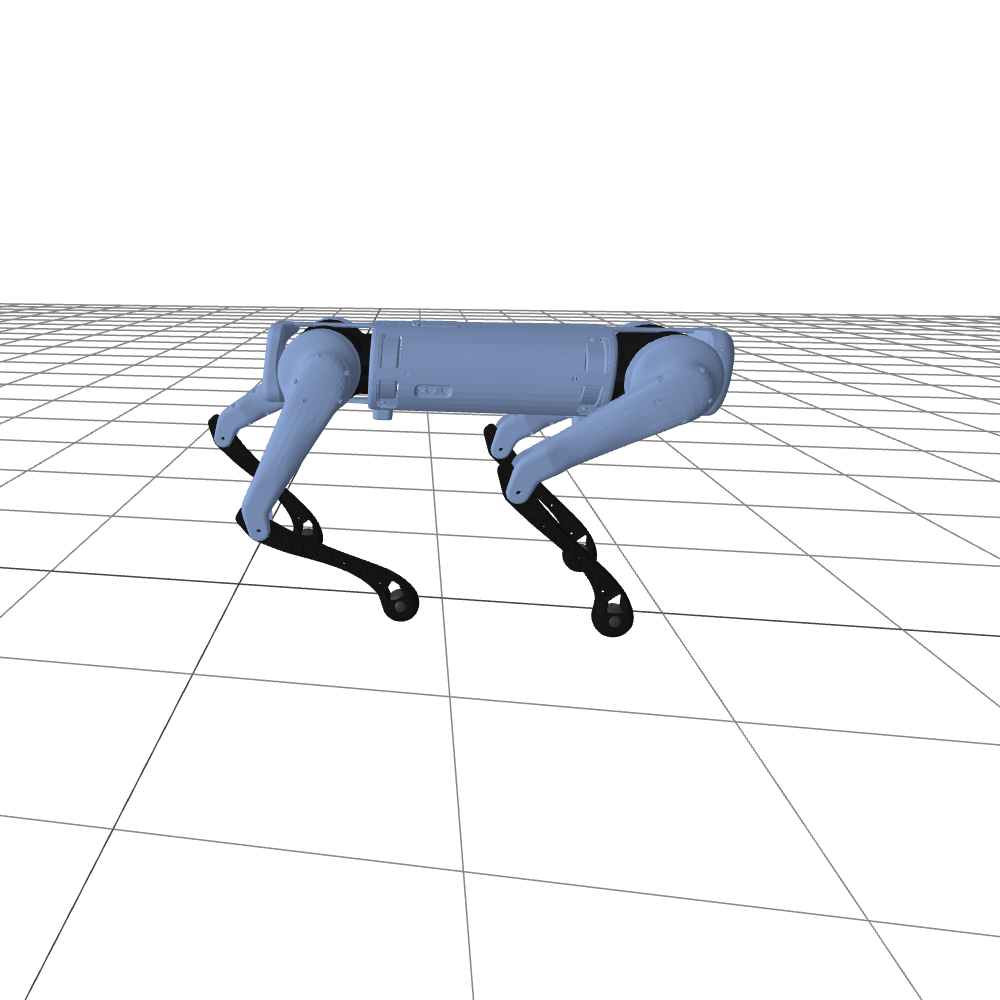}%
            \hfill%
            \includegraphics[width=0.32\linewidth]{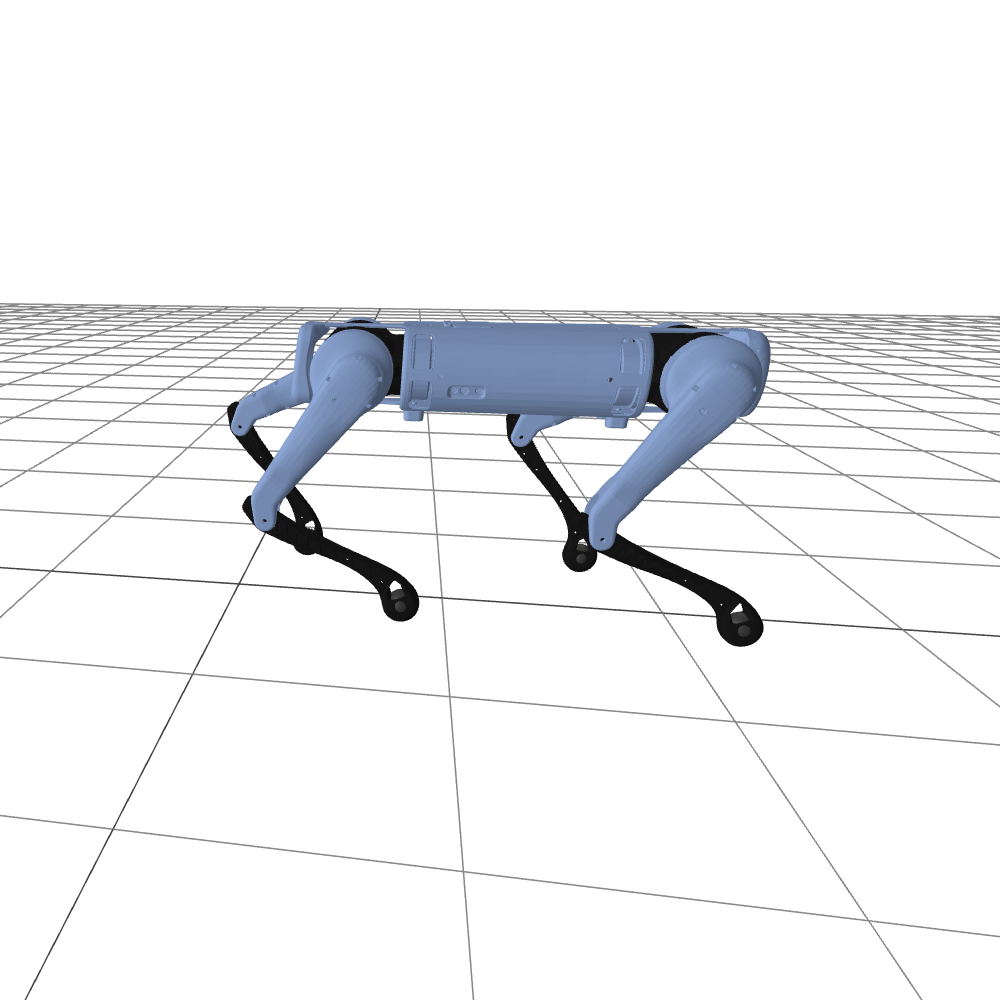}%
            \hfill%
            \includegraphics[width=0.32\linewidth]{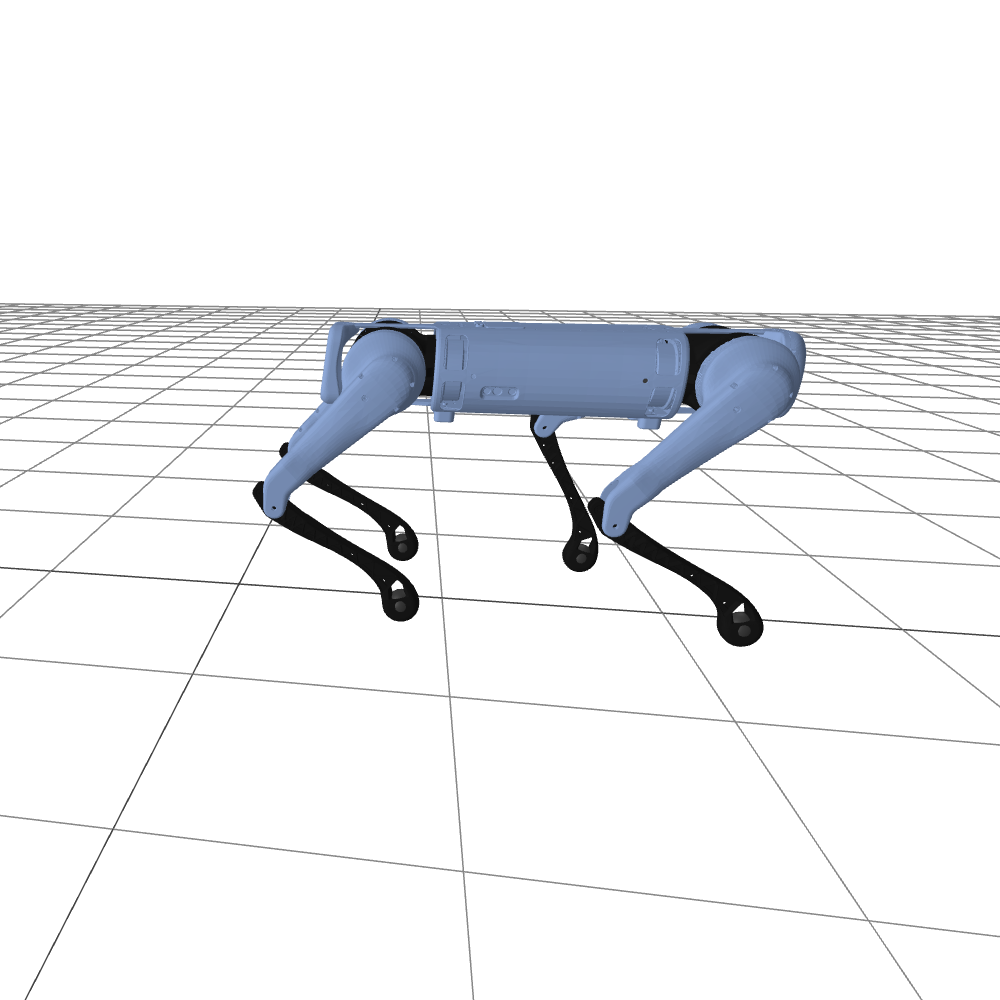}
        \end{minipage}
        \label{fig:b1_motion}
    }
    \par\smallskip
    \subfloat[CoM tracking]{
        \includegraphics[width=0.98\columnwidth]{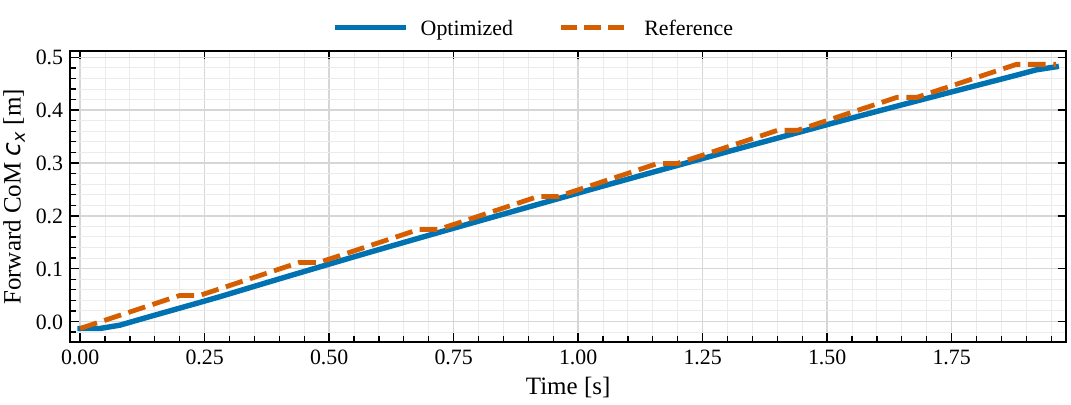}
        \label{fig:b1_com}
    }
    \par\smallskip
    \subfloat[Contact forces]{
        \includegraphics[width=0.98\columnwidth]{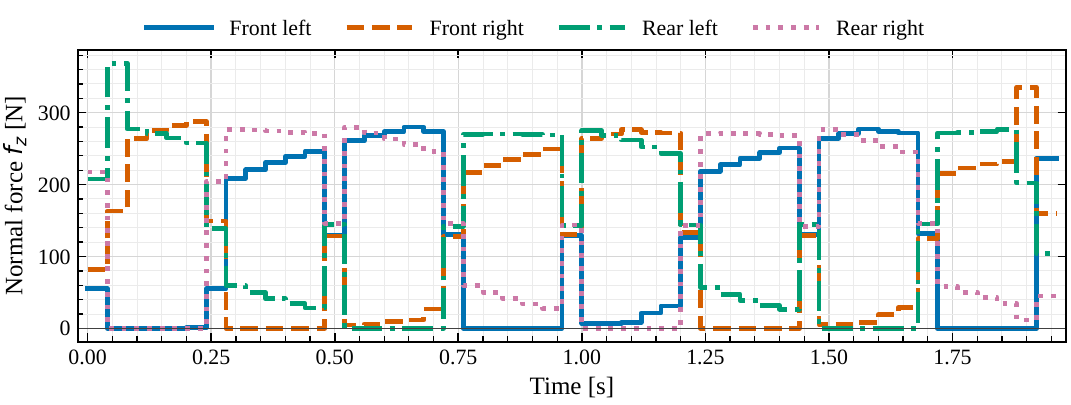}
        \label{fig:b1_forces}
    }
    \caption{Unitree B1 walking using SLP with HALO-LP.
    (a) Snapshots of the optimized motion.
    (b) Optimized and reference forward CoM position.
    (c) Vertical contact forces at the four feet.}
    \label{fig:b1_walking}
\end{figure}

\begin{figure}[!t]
    \centering
    \subfloat[Trotting motion]{
        \begin{minipage}{0.98\columnwidth}
            \centering
            \includegraphics[width=0.32\linewidth]{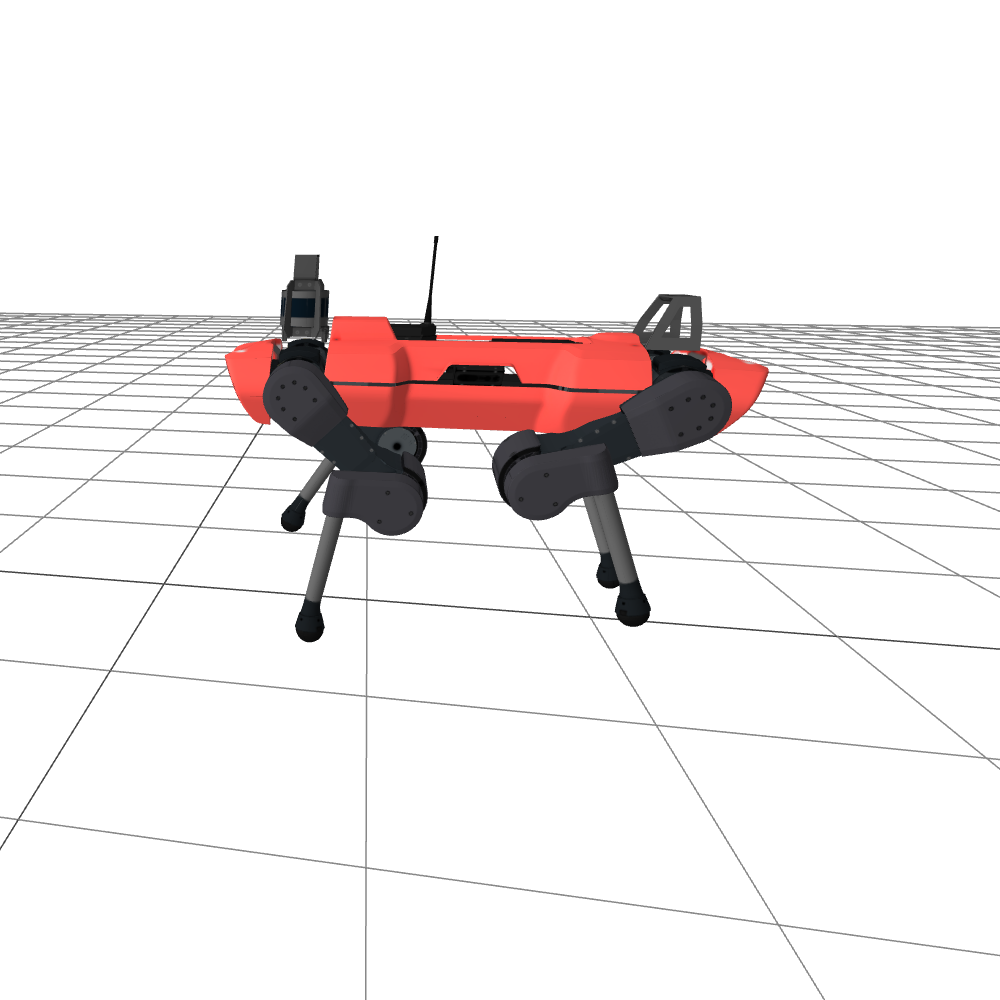}%
            \hfill%
            \includegraphics[width=0.32\linewidth]{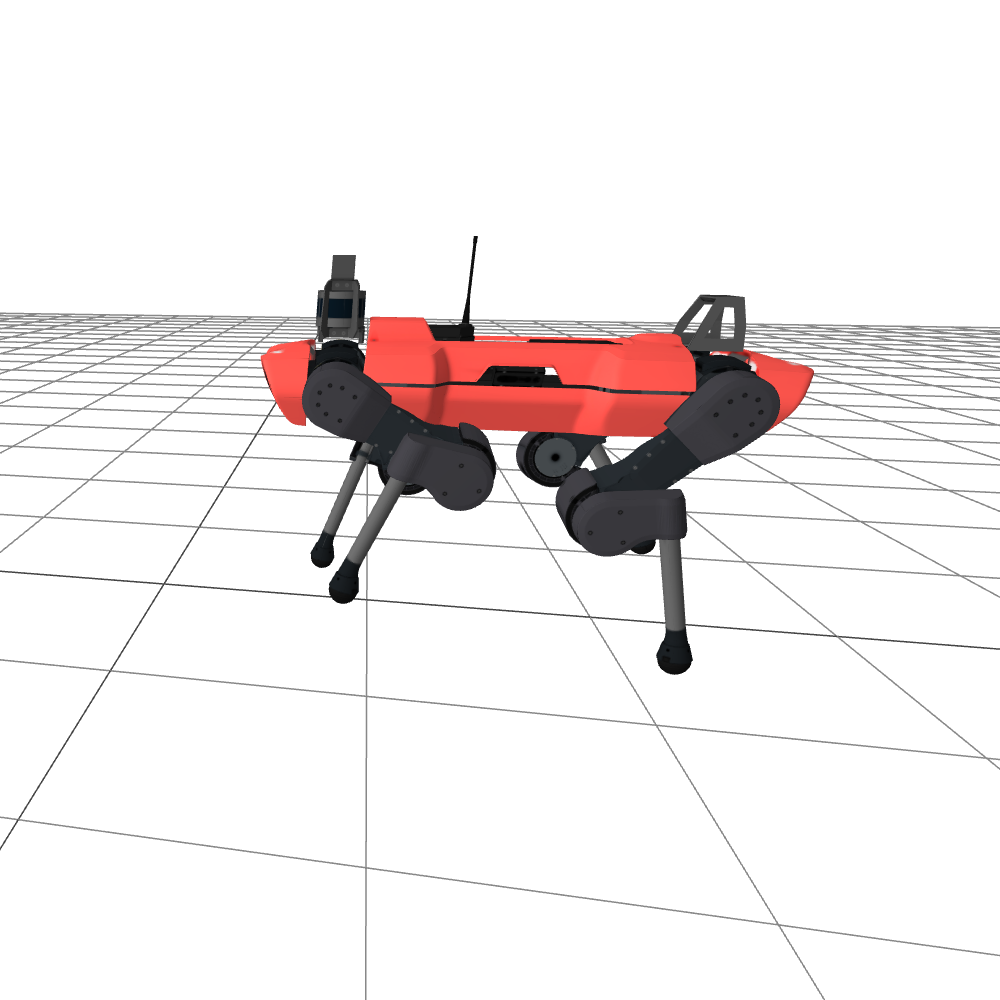}%
            \hfill%
            \includegraphics[width=0.32\linewidth]{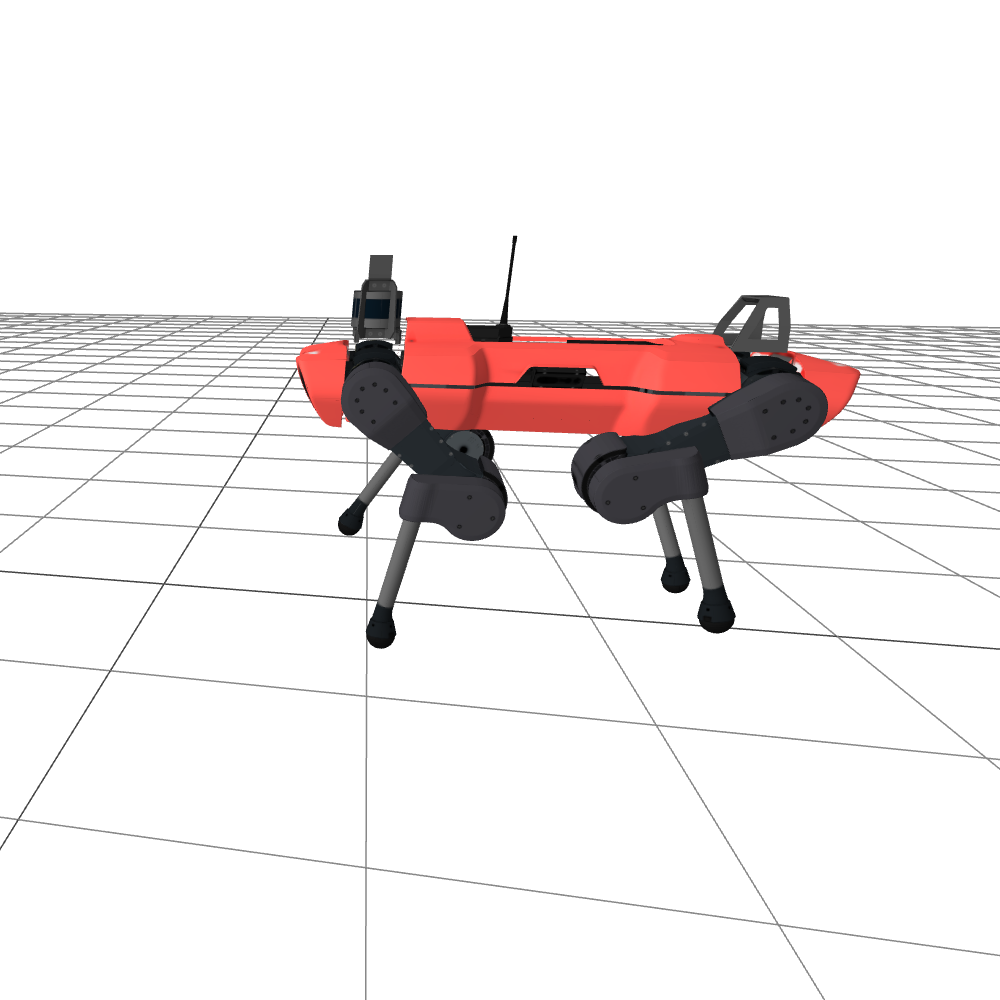}
        \end{minipage}
        \label{fig:anymal_motion}
    }
    \par\smallskip
    \subfloat[CoM tracking]{
        \includegraphics[width=0.98\columnwidth]{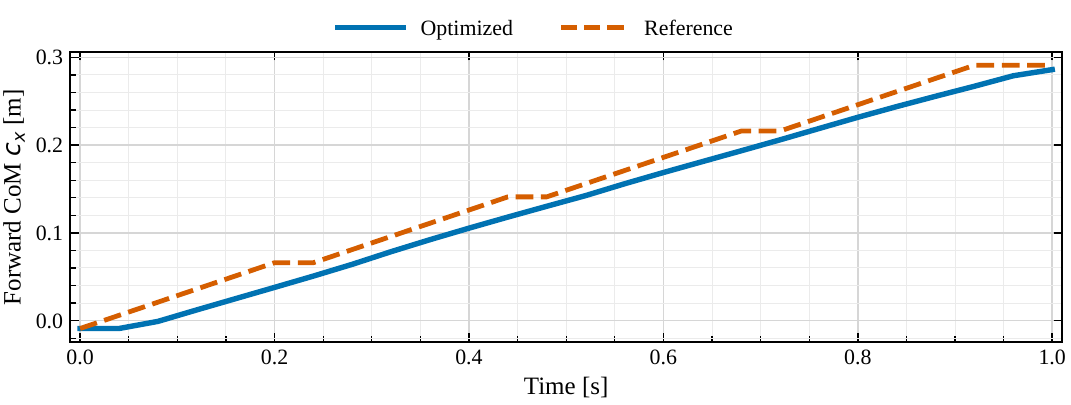}
        \label{fig:anymal_com}
    }
    \par\smallskip
    \subfloat[Contact forces]{
        \includegraphics[width=0.98\columnwidth]{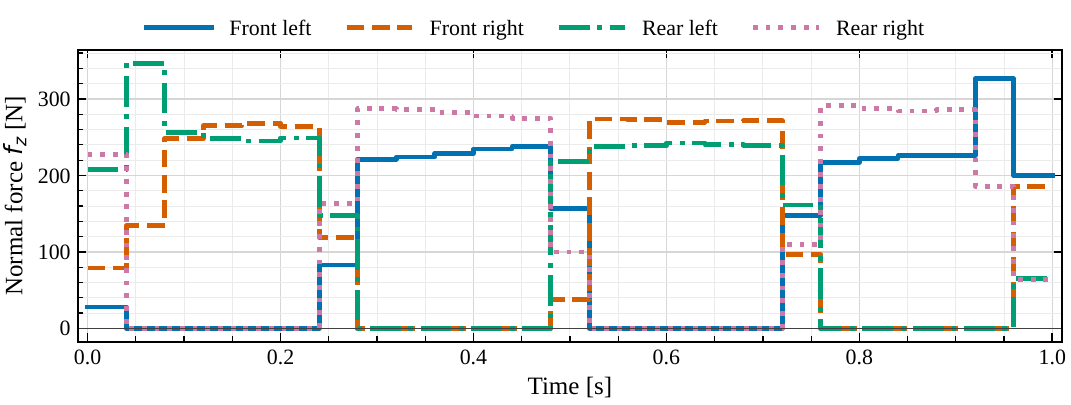}
        \label{fig:anymal_forces}
    }
    \caption{ANYmal-C trotting using SLP with HALO-LP.
    (a) Snapshots of the optimized motion.
    (b) Optimized and reference forward CoM position.
    (c) Vertical contact forces at the four feet.}
    \label{fig:anymal_trotting}
\end{figure}

We consider walking and trotting with fixed contact
schedules, a discretization step of $0.04$\,s, and
a friction coefficient of $0.7$. The respective horizons
contain 49 and 25 shooting intervals. Figs.~\ref{fig:b1_walking} and~\ref{fig:anymal_trotting}
show the walking and trotting motions obtained using SLP
with HALO-LP for the Unitree B1 and ANYmal-C quadrupeds,
respectively. In both experiments, the optimized forward
CoM position closely tracks its reference, while the normal
contact forces reflect the prescribed gait patterns, with
vanishing forces at inactive contacts.

\section{Conclusion}
This letter presented HALO-LP, a homogeneous all-shifted
non-interior-point method for linear programming that
combines effective warm-starting with primal and dual
infeasibility detection. By relaxing strict interiority
requirements while preserving the homogeneous self-dual
structure, HALO-LP provides a unified framework for
solving sequences of related LPs. Numerical experiments
demonstrated superior warm-starting performance on the
evaluated Netlib instances, successful
optimality and infeasibility detection, and applicability
to quadrupedal optimal control through sequential linear
programming.

Future work will extend the homogeneous all-shifted
framework to second-order cone programming (SOCP)
and develop structure-exploiting linear solvers
for optimal control based on Riccati recursions
and their parallel implementations. We also aim
to investigate its integration into mixed-integer
convex optimization, leveraging warm-starting and
infeasibility certificates to efficiently solve
continuous relaxations.

\newpage

\vfill

\end{document}